\documentclass{article}

\usepackage{microtype}
\usepackage{graphicx}
\usepackage{subfigure}
\usepackage{booktabs} % for professional tables

\usepackage{hyperref}

\usepackage[accepted]{icml2024}

\usepackage{amsmath}
\usepackage{amssymb}
\usepackage{mathtools}
\usepackage{amsthm}

\usepackage[capitalize,noabbrev]{cleveref}

\theoremstyle{plain}
\newtheorem{theorem}{Theorem}[section]
\newtheorem{proposition}[theorem]{Proposition}

\theoremstyle{definition}

\theoremstyle{remark}

\usepackage[textsize=tiny]{todonotes}

\DeclareMathOperator*{\argmax}{arg\,max}
\DeclareMathOperator*{\argmin}{arg\,min}

\icmltitlerunning{Optimistic Multi-Agent Policy Gradient}

\begin{document}

\twocolumn[
\icmltitle{Optimistic Multi-Agent Policy Gradient}




\begin{icmlauthorlist}
\icmlauthor{Wenshuai Zhao}{aalto_ee}
\icmlauthor{Yi Zhao}{aalto_ee}
\icmlauthor{Zhiyuan Li}{uestc}
\icmlauthor{Juho Kannala}{aalto_cs,oulu}
\icmlauthor{Joni Pajarinen}{aalto_ee}
\end{icmlauthorlist}

\icmlaffiliation{aalto_ee}{Department of Electrical Engineering and Automation, Aalto University, Finland}
\icmlaffiliation{aalto_cs}{Department of Computer Science, Aalto University, Finland}
\icmlaffiliation{uestc}{School of Computer Science and Engineering, University of Electronic Science and Technology of China, China}
\icmlaffiliation{oulu}{University of Oulu, Finland}

\icmlcorrespondingauthor{Wenshuai Zhao}{wenshuai.zhao@aalto.fi}

\icmlkeywords{Multi-agent reinforcement learning, Relative overgeneralization, Optimism}

\vskip 0.3in
]



\printAffiliationsAndNotice{}  % leave blank if no need to mention equal contribution

\begin{abstract}
\textit{Relative overgeneralization} (RO) occurs in cooperative multi-agent learning tasks when agents converge towards a suboptimal joint policy due to overfitting to suboptimal behaviors of other agents. No methods have been proposed for addressing RO in multi-agent policy gradient (MAPG) methods although these methods produce state-of-the-art results. To address this gap, we propose a general, yet simple, framework to enable optimistic updates in MAPG methods that alleviate the RO problem. Our approach involves clipping the advantage to eliminate negative values, thereby facilitating optimistic updates in MAPG. The optimism prevents individual agents from quickly converging to a local optimum. Additionally, we provide a formal analysis to show that the proposed method retains optimality at a fixed point. In extensive evaluations on a diverse set of tasks including the \textit{Multi-agent MuJoCo} and \textit{Overcooked} benchmarks, our method outperforms strong baselines on 13 out of 19 tested tasks and matches the performance on the rest.
\end{abstract}

\section{Introduction}

Multi-agent reinforcement learning (MARL) is a promising approach for many cooperative multi-agent decision making applications, such as those found in robotics and wireless networking~\cite{busoniu2008comprehensive}. However, despite recent success on increasingly complex tasks~\cite{vinyals2019grandmaster, yu2022surprising}, these methods can still fail on simple two-player matrix games as shown in Figure~\ref{fig:matrix}. The underlying problem is that in cooperative tasks, agents may converge to a suboptimal joint policy when updating their individual policies based on data generated by
other agents' policies which have not converged yet. This phenomenon is called \textit{relative overgeneralization}~\cite{wiegand2004analysis} (RO) and has been widely studied in tabular matrix games~\cite{claus1998dynamics, lauer2000algorithm, panait2006lenient}
but remains an open problem in state-of-the-art MARL methods~\cite{de2020independent, yu2022surprising, kuba2021trust}.
% ... \joni{General motivation needed here. There are also important application domains such as robotics and wireless networking.}

The cause for RO can be understood intuitively. In a cooperative multi-agent system, the common reward derives from the joint actions. From the perspective of a single agent, an optimal individual action may still incur low joint reward due to non-cooperative behaviors of other agents. This is common in cooperative tasks when iteratively optimizing individual policies, especially at the beginning of training since individual agents have not learned to act properly to cooperate with others. It is particularly challenging in tasks with a large penalty for incorrect joint actions, such as the pitfalls in the \textit{climbing} matrix game in Figure~\ref{fig:matrix} leading often to agents that prefer a suboptimal joint policy. 

Existing techniques to overcome the pathology share the idea of applying optimistic updating in Q-learning with different strategies to control the degree of optimism~\cite{lauer2000algorithm} such as in hysteretic Q-learning~\cite{matignon2007hysteretic} or in lenient agents~\cite{panait2006lenient}. However, these methods are designed based on Q-learning and only tested on matrix games~\cite{matignon2012independent} with tabular Q representation. Although optimism successfully helps overcome \textit{RO} problem and converges to global optima in tabular tasks, it could amplify the overestimation problem when combined with DQN~\cite{van2016deep} with function approximation~\cite{omidshafiei2017deep, palmer2017lenient, rashid2020weighted}, as verified in our experiments. On the other hand, recent multi-agent policy gradient (MAPG) methods have achieved state-of-the-art performance on popular MARL benchmarks~\cite{de2020independent, yu2022surprising, sun2023trust, wang2023order} but still suffer from the \textit{RO} problem, converging to a low value local optimum in simple matrix games. The above drawbacks motivate this work to investigate whether optimism can be applied to MAPG methods and whether it can boost performance further by mitigating the \textit{RO} problem. 

Our contribution is threefold: (1) To our knowledge, we are the first to investigate the application of optimism in MAPG methods. We propose a general, yet simple, framework to incorporate optimism into the policy gradient computation in MAPG. Specifically, we propose to clip the advantage values $A(s, a^i)$ when updating the policy. For completeness, we extend the framework to include a hyperparameter to control the degree of optimism, resulting in a \textit{Leaky ReLU} function~\cite{maas2013rectifier} to reshape the advantage values. (2) We provide a form analysis to show that the proposed method retains optimality at a fixed point. (3) In our experiments\footnote{Source Code: \href{https://github.com/wenshuaizhao/optimappo}{https://github.com/wenshuaizhao/optimappo}}, the proposed OptiMAPPO algorithm successfully learns global optima in matrix games and outperforms both recent state-of-the-art MAPG methods MAPPO~\cite{yu2022surprising}, HAPPO, HATRPO~\cite{kuba2021trust}, and existing optimistic methods in complex domains.

\section{Related work}

% \joni{We should add a short general overview of related work here that motivates the related work below starting from our contribution.}

In this section, we discuss classic optimistic methods and optimistic DQN based approaches. We also discuss recent MAPG methods that yield state-of-the-art performance on common benchmarks. Optimistic Thompson sampling (O-TS)~\cite{hu2023optimistic} is also discussed as it utilizes a similar clipping technique to improve exploration for stochastic bandits. General multi-agent exploration methods are related and introduced. For completeness, we further discuss advantage shaping in single-agent settings. 
% \textit{Relative overgeneralization}~\cite{wiegand2004analysis} has been extensively studied in repeated games. 

% \wenshuai{Done}
\paragraph{Classic Optimistic Methods} To the best of our knowledge, distributed Q-learning~\cite{lauer2000algorithm} proposes the first optimistic updating in independent Q-learning, where $Q$-values are only updated when the value increases. \citet{lauer2000algorithm}  also provides brief proof that the proposed optimism-based method converges to the global optimum in deterministic environments. To handle stochastic settings, hysteretic Q-learning~\cite{matignon2007hysteretic} adjusts the degree of optimism by setting different learning rates for the Q values of actions with different rewards. The frequency maximum Q value heuristic (FMQ)~\cite{kapetanakis2002reinforcement} considers changing the action selection strategies during exploration instead of modifying the updating of Q values. Lenient agent~\cite{panait2006lenient, wei2016lenient} employs more heuristics to finetune the degree of optimism by initially adopting an optimistic disposition and gradually transforming into average reward learners.

\paragraph{Optimistic Deep Q-Learning}
% \joni{This paragraph should be shortened.}
 Recently, several works have extended optimistic methods and lenient agents to Deep Q-learning (DQN). Dec-HDRQN~\cite{omidshafiei2017deep} and lenient-DQN~\cite{palmer2017lenient} apply hysteretic Q-learning 
 and leniency to DQN, respectively. Optimistic methods have also been combined with value decomposition methods. Weighted QMIX~\cite{rashid2020weighted} uses a higher weight to update the Q-values of joint actions with high rewards and a lower weight to update the values of suboptimal actions. FACMAC~\cite{peng2021facmac} improves MADDPG~\cite{lowe2017multi} by taking actions from other agents' newest policies while computing the Q values, which can also be regarded as optimistic updating as it is natural to suppose the newer policies of other agents would generate better joint Q-values. Even though the optimism has been applied to DQN-based methods, in common benchmarks they are usually outperformed by recent MAPG methods~\cite{yu2022surprising}. This unsatisfying performance of optimistic DQN methods can be attributed to the side effect of overestimation of Q-values, which we also empirically verify in our experiments.

 \paragraph{Multi-Agent Policy Gradient Methods}
 COMA~\cite{foerster2018counterfactual} is an early MAPG method in parallel with value decomposition methods~\cite{sunehag2017value, rashid2020monotonic}. It learns a centralized on-policy $Q$ function and utilizes it to compute the individual advantage for each agent to update the policy. However, until IPPO and MAPPO~\cite{de2020independent, yu2022surprising} which directly apply single-agent PPO~\cite{schulman2017proximal} into multi-agent learning, MAPG methods show significant success on popular benchmarks. The strong performance of these methods might be credited to the property that individual trust region constraint in IPPO/MAPPO can still lead to a centralized trust region as in single-agent PPO~\cite{sun2023trust}. Nonetheless, HAPPO/HATRPO~\cite{kuba2021trust} and A2PO~\cite{wang2023order} further improve the monotonic improvement bound by enforcing joint and individual trust region, while with the cost of sequentially updating each agent. However, even with a strong performance on popular benchmarks, these methods don't explicitly consider the \textit{relative overgeneralization} problem during multi-agent learning and can still converge to a suboptimal joint policy.  

 \paragraph{Multi-Agent Exploration Methods}

 Since RO can be seen as a special case of the general exploration problem we discuss below general MARL exploration methods. Similar to single-agent exploration, ~\citet{rashid2020monotonic} and \citet{hu2021policy} use noise for exploration in multi-agent learning. However, as shown by~\cite{mahajan2019maven} noise-based exploration can result in suboptimal policies. For coordinated exploration, multi-agent variational exploration~\cite{mahajan2019maven} conditions the joint Q-value on a latent state. \citet{jaques2019social} maximizes the mutual information between agent behaviors for coordination. This may nevertheless still lead to a sub-optimal joint strategy~\cite{li2022pmic}. \citet{zhao2023conditionally} propose conditionally optimistic exploration (COE) which augments agents’ Q-values by an optimistic bonus based on a global state-action visitation count of preceding agents. However, COE is designed for discrete states and actions and is difficult to scale up to complex tasks with continuous state and action spaces. Cooperative multi-agent exploration (CMAE)~\cite{liu2021cooperative} only counts the visitations of states in a restricted state space to learn an additional exploration policy. However, the restricted space selection is hard to scale up and thus CMAE fails to show performance improvement on widely used MARL benchmarks. Our method aims to mitigate the specific RO problem with state-of-the-art MAPG methods in order to further boost their performance on complex tasks.

\paragraph{Optimistic Thompson Sampling} Thompson sampling is a popular method for stochastic bandits~\cite{russo2018tutorial}. Conceptually, Thompson Sampling plays an action according to the posterior probability distribution of the optimal action. The key
idea for optimistic Thompson sampling (O-TS)~\cite{hu2023optimistic} is to clip the posterior distribution in an optimistic way to ensure that the sampled models are always better than the empirical models~\cite{chapelle2011empirical}. O-TS shares similar heuristics with our method, that is, there is no need to decrease a prediction in Thompson sampling, or no need to explicitly decrease the action probability in policy updates. However, it is not straightforward to apply O-TS to model-free policy gradient methods. Our method achieves optimism by a novel advantage clipping instead of the posterior distribution clipping in O-TS.
 
\paragraph{Advantage Shaping}
% \wenshuai{This part needs to be edited more}
A few works have explored advantage shaping in single-agent RL settings. PPO-CMA~\cite{hamalainen2020ppo} tackles the prematurely shrinking variance problem in PPO by clipping or mirroring the negative advantages in order to increase exploration and eventually converges to a better policy. Self-imitation learning (SIL)~\cite{oh2018self} learns an off-policy gradient from replay buffer data with positive advantages in addition to the regular on-policy gradient. However, different from these existing works, we are motivated by solving the \textit{RO} problem in multi-agent learning.

\section{Background}\label{sec: preliminary}
We begin by introducing our problem formulation. Following this, we present the concept of optimistic Q-learning as it forms the basis of hysteresis and leniency based approaches which differ in their ways regulating the degree of optimism. 

\subsection{Problem Formulation}
We mainly study the fully cooperative multi-agent sequential decision-making tasks which can be formulated as a \textit{multi-agent Markov decision process} (MMDP)~\cite{boutilier1996planning} consisting of a tuple $(\mathcal{S}, \{\mathcal{A}^i\}_{i \in \mathcal{N}}, r, \mathcal{P}, \gamma)$, where $\mathcal{N}=\{1, \cdots, n\}$ is the set of agents. At time step $t$ of Dec-MDP, each agent $i$ observes the full state $s_t$ in the state space $\mathcal{S}$ of the environment, performs an action $a^i_t$ in the action space $\mathcal{A}^i$ from its policy $\pi^i(\cdot \vert s_t)$. The joint policy consists of all the individual policies $\mathbf{\pi}(\cdot \vert s_t)=\pi^1\times \cdots \times \pi^n$. The environment takes the joint action of all agents $\mathbf{a}_t=\{a_t^1, \cdots, a_t^n\}$, changes its state following the dynamics function $\mathcal{P}: \mathcal{S}\times \mathcal{A} \times \mathcal{S} \mapsto [0, 1]$ and generates a common reward $r: \mathcal{S}\times \mathcal{A} \mapsto \mathbb{R}$ for all the agents. $\gamma \in [0, 1)$ is a reward discount factor. The agents learn their individual policies and maximize the expected return: $\mathbf{\pi}^{\ast}=\argmax_{\mathbf{\pi}}\mathbb{E}_{s,\mathbf{a}\sim\mathbf{\pi}, \mathcal{P}}[\sum_{t=0}^{\infty}\gamma^{t}r(s_t, \mathbf{a}_t)]$. Since we train independent policy networks for each agent, we use $s$ and $s^i$ interchangeably as they all include the full state information in our experiments.

\subsection{Optimistic Q-learning}
We explain the idea of optimistic Q-learning~\cite{matignon2007hysteretic} based on hysteretic Q-learning. While regular Q-learning update assigns the same learning rate to both negative and positive updates, hysteretic Q-learning assigns a higher weight to the positive update of the Q value, i.e., when the right-hand side (RHS) of Equation~\ref{equ:q_update} has a higher value than the left-hand side (LHS). In our experiments on the baseline hysteretic Q-learning, it is equivalent to only set the weight for negative updates, leaving the positive update with the default learning rate. 
\begin{equation}
\begin{aligned}
    % \vskip -0.15in
    Q(s_t, a_t)\leftarrow &Q(s_t, a_t)+ \\
    &\alpha [r+\gamma \max_{a}Q(s_{t+1}, a)-Q(s_t, a_t)]
\end{aligned}
\label{equ:q_update}
\end{equation}

Optimism with tabular Q-learning has been demonstrated effective to solve the \textit{RO} problem. However, with function approximation, the optimistic update could exacerbate the overestimation problem~\cite{van2016deep} of deep Q-learning and thus fail to improve the underlying methods, which is also shown in our experiments. To our knowledge, the application of optimism in MAPG methods has not been explored and it remains unclear how to facilitate optimism in policy gradient methods and how much improvement it can escort.

\section{Method}\label{sec: method}
Aware of the importance of optimism in solving \textit{RO} problem and the limitation of optimistic Q-learning, we instead propose a principled way to apply optimism to the recent MAPG methods. The algorithm we use in experiments is instantiated based on MAPPO~\cite{yu2022surprising} but note that the proposed framework can be further applied to other advantage actor-critic (A2C) based MARL methods as shown in Appendix~\ref{sec:maa2c}.

\subsection{Optimistic MAPPO}
In MAPPO~\cite{yu2022surprising}, each agent learns a centralized state value function $V(s)$, and the individual policy is updated via maximizing the following objective
% \begin{equation}
%     \max_{\pi_{\theta^i}}\mathbb{E}_{(s^i, a^i)\sim \pi^i}[\min(r(\theta)A(s^i, a^i), \text{clip}(r(\theta), 1\pm \epsilon)A(s^i, a^i))],
% \end{equation}

\begin{equation}
    \begin{aligned}
    \max_{\pi_{\theta^i}}\mathbb{E}_{(s^i, a^i)\sim \pi^i}[&\min(r(\theta)A(s^i, a^i), \\
    &\text{clip}(r(\theta), 1\pm \epsilon)A(s^i, a^i))],
    \end{aligned}
\end{equation}

where $\epsilon$ is the clipping threshold and $r(\theta)$ is the importance ratio between the current policy and the previous policy used to generate the data,
\begin{equation}
    r(\theta)=\frac{\pi_{\theta^i}(a^i_t\vert s^i_t)}{\pi_{\theta^i_{\text{old}}}(a^i_t\vert s^i_t)}
\end{equation}

The advantage $A(s^i, a^i)$ is usually estimated by the generalized advantage estimator (GAE)~\cite{schulman2015high} defined as
\begin{equation}
    A_t^{\text{GAE}(\lambda, \gamma)}=\sum_{l=0}^{\infty}(\gamma\lambda)^l\delta_{t+l},
    \label{equ:gae}
\end{equation}
and $\delta_{t}$ denotes the TD error
\begin{equation}
    \delta_t=r_t+\gamma V(s_{t+1})-V(s_t)
    \label{equ:td}
\end{equation}

While PPO~\cite{schulman2017proximal} adopts the clipping operation to constrain the policy change in order to obtain the guaranteed monotonic improvement, the policy update is similar to common A2C~\cite{mnih2016asynchronous} methods. The policy is improved by increasing the actions with positive advantage values and decreasing the others with negative advantages. However, the actions currently with negative advantages might be the optimal action and the current negation comes from the currently suboptimal teammates, not from the suboptimality of the actions. In tasks with a severe \textit{relative overgeneralization} problem, optimal actions are often not recovered by simple exploration strategies and the joint policy converges to a suboptimal solution.

In order to overcome the \textit{relative overgeneralization} problem in MAPG methods, the proposed \textit{optimistic MAPPO} (OptiMAPPO) applies a clipping operation to reshape the estimated advantages~\cite{schulman2015high}. OptiMAPPO optimizes the agents' policies by maximizing the following new objective

\begin{equation}
\begin{aligned}
    \max_{\pi_{\theta^i}}\mathbb{E}_{(s^i, a^i)\sim \pi^i}[&\min(r(\theta)\text{clip}(A(s^i, a^i), 0), \\
    &\text{clip}(r(\theta), 1\pm \epsilon)\text{clip}(A(s^i, a^i), 0))],
\end{aligned}
\label{equ:policy_update}
\end{equation}

where $\text{clip}(A(s^i, a^i), 0)$ denotes that negative advantage estimates are clipped to zero while positive advantage values remain unchanged. $\text{clip}(r(\theta), 1\pm \epsilon)$ is the same clipping operation in PPO~\cite{schulman2017proximal}. The proposed advantage clipping operation allows to be optimistic to temporarily suboptimal actions incurred by the \textit{RO} problem and facilitates individual agents to converge to a better joint policy. The implementation can be as straightforward as a single-line modification of underlying MAPG methods. However, as demonstrated by our experiments, the effectiveness of optimism is significantly enhanced compared to optimistic Q-learning based methods.

\paragraph{Extension to \textit{Leaky ReLU} operation} The proposed clipping operation can be seen as a special case of a \textit{Leaky ReLU} (LR) operation of advantage values where there is a hyperparameter $\eta\in [0,1]$ to control the degree of optimism, $\text{LR}(A)=\max(\eta A, A)$. Our clipping operation is the case when $\eta=0$, while $\eta=1$ recovers the original MAPG methods. In our experiments, we find that the performance is improved more while setting lower $\eta$, i.e. higher optimism. However, we argue that such an extension could be beneficial in stochastic reward environments as discussed in hysteretic Q-learning~\cite{matignon2007hysteretic} and lenient agents methods~\cite{palmer2017lenient, matignon2012independent}. The extension allows us to control the degree of optimism in a finer granularity. We leave the study in stochastic environments as future work and this paper is primarily focused on the first to investigate the effectiveness of optimism in MAPG methods.

\subsection{Analysis}
The following analysis shows formally from an operator view that the proposed algorithm retains optimality at a fixed point.

The operator view fits our method in the context of policy gradient methods. As shown in~\cite{ghosh2020operator}, the policy update in policy gradient methods can be seen as two successive operations from \textit{improvement operator} $\mathcal{I}_V$ and \textit{projection operator} $\mathcal{P}_V$. Such an operator view connects both Q-learning and vanilla policy gradient by accounting to different $\mathcal{I}_V$, which is detailed in the Appendix~\ref{sec:operator_intro}. 

We show that the proposed advantage clipping forms a new \textit{improvement operator} that retains optimality at a fixed point of the operators. To simplify the analysis, we only take the clipping operation instead of the extended \textit{Leaky ReLU} operation, as it transforms the advantage values into non-negative values, satisfying the valid probability distribution requirement in the operator view derivation. Based on the clipped advantage, the new policy gradient is
\begin{equation}\label{equ: lr_gradient}
\begin{aligned}
    \theta_{t+1}=&\theta_t + \epsilon\sum_s d^{\pi_t}(s)\sum_a \pi_t (a\vert s)\cdot\\
    &\text{clip}(A^{\pi_t}(s,a))\frac{\partial\log \pi_{\theta}(a \vert s)}{\partial \theta}\mid_{\theta=\theta_t}.
\end{aligned}
\end{equation}
Note that we use $\text{clip}(A^{\pi_t}(s,a))$ and $\text{clip}(A^{\pi_t}(s,a), 0)$ interchangeably. It corresponds to the new \textit{improvement operator} $\mathcal{I}_V^{\text{clip}}$ formulated as
% We are interested in seeking a better $\mathcal{I}_V$ by transforming $Q(s, a)$ in policy gradient. Fitting our proposed advantage shaping into the policy gradient, it can be understood as a new algorithm using $\text{LR}(A(s, a))+V(s)$ instead of the $Q(s, a)= A(s, a)+ V(s)$ in Equation~\ref{equ: improve}. The new \textit{improvement operator} $\mathcal{I}_V^{\text{LR}}$ will be,
\begin{equation}
    \mathcal{I}_V^{\text{clip}} \pi(s,a)=(\frac{1}{\mathbb{E}_{\pi}[V^{+\pi}]}d^{\pi}(s)V^{+\pi}(s))\text{clip}(A^{\pi})\pi(a\vert s),
\end{equation}
where $\text{clip}(A^{\pi})\pi$ and $V^{+\pi}$ are defined as:
\begin{equation}
\begin{aligned}
    &\text{clip}(A^{\pi})\pi(a\vert s)=\frac{1}{V^{+\pi}(s)}\text{clip}(A^{\pi}(s,a))\pi(a\vert s),\\
    &V^{+\pi}(s)=\sum_a \text{clip}(A(s,a))\pi(a\vert s).
\end{aligned}
\end{equation}

The \textit{projection operator} remains the same as $\mathcal{P}_{V\mu}$ in vanilla policy gradient. With the clipped advantage, the optimal policy $\pi^{\ast}(a\vert s)$ is a fixed point of the operators $\mathcal{I}_V^{\text{clip}}\circ \mathcal{P}_V$ as in the vanilla policy gradient. The property is shown in the following Proposition~\ref{thm:thm2} and proven in the Appendix~\ref{sec:proof}.
% ~\ref{proof:operator}.

\begin{proposition}\label{thm:thm2}
$\pi(\theta^{\ast})$ is a fixed point of $\mathcal{I}_V^{\text{clip}}\circ \mathcal{P}_V$,
\end{proposition}
where $\circ$ denotes the function composition notation. It means the latter operator $\mathcal{P}_V$ is first evaluated and then its output will be used by the former operator $\mathcal{I}_V^{clip}$.

\begin{figure*}[ht]
\centering
    \includegraphics[width=0.19\textwidth]{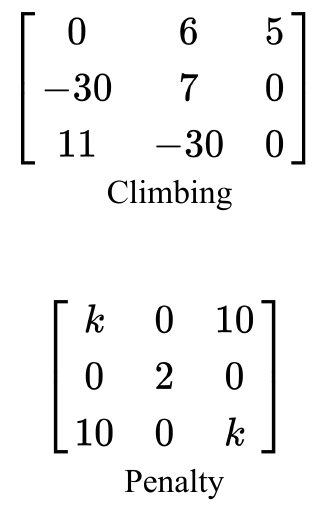}
    \hfill
    \vrule height 150pt depth -12pt width 0.5pt
    \hfill
    \includegraphics[trim= 90 30 0 30,clip,width=0.8\textwidth]{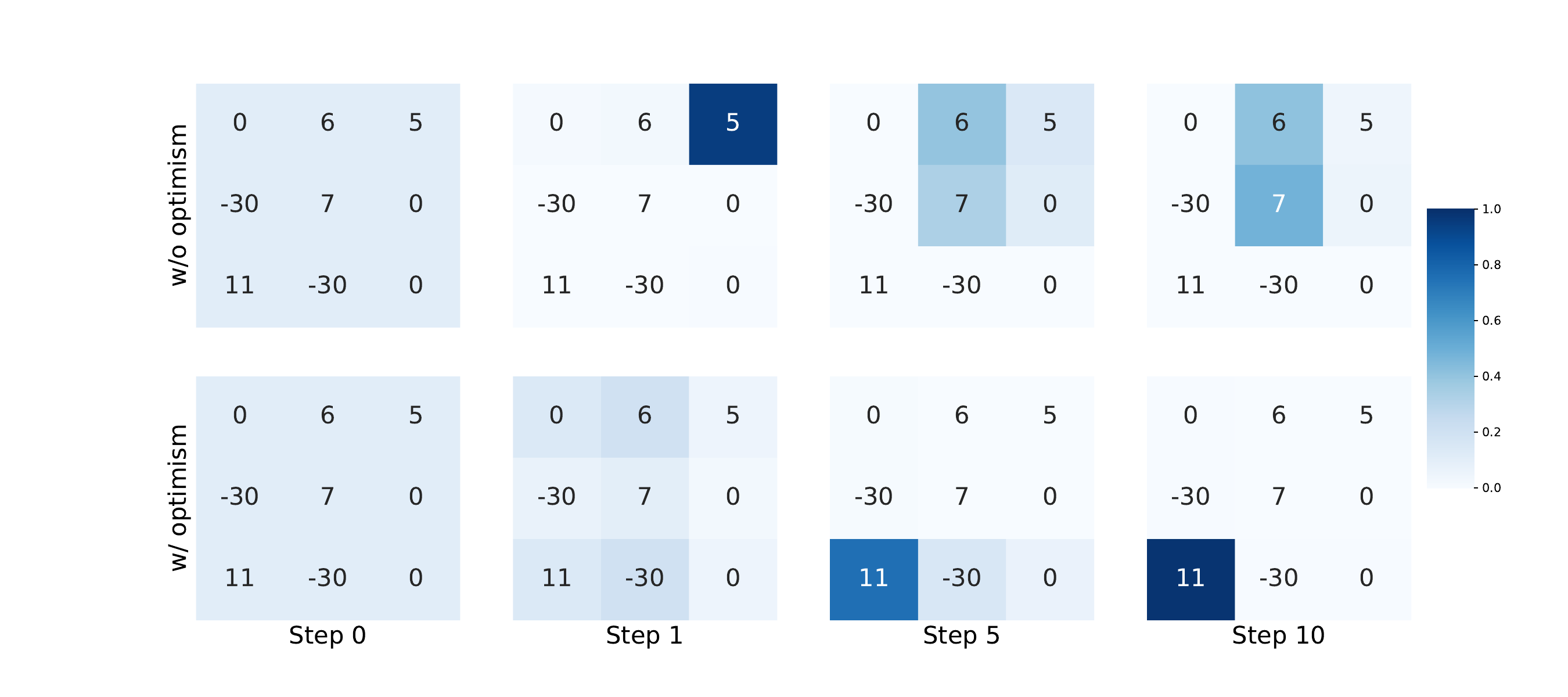}
  \caption{\textbf{Left:} Payoff matrix of the \textit{climbing} and \textit{penalty}. Each game has two agents, which select the row and column index respectively to find the maximal element of the matrix. \textbf{Right:} The comparison of the learning process with and without an optimistic update on the Climbing task. It shows that the optimistic update is necessary to solve the RO problem.}
\label{fig:matrix}
\end{figure*}

\section{Experiments}

We compare our method with the following strong MAPG baselines on both illustrative matrix games and complex domains including \textit{Multi-agent MuJoCo} and \textit{Overcooked}:
% \footnote{Noticed that we do not choose value-based methods as our baselines since most of them are outperformed by MAPPO.}:
\begin{itemize}
    \item \textbf{MAPPO} directly applies singe-agent PPO into multi-agent settings while with a centralized critic. Despite the lack of theoretical guarantee, MAPPO has achieved tremendous success in a variety of popular benchmarks.
    \item \textbf{HATRPO} is currently one of the SOTA
    MAPG algorithms that leverages \textit{Multi-Agent Advantage Decomposition Theorem}~\cite{kuba2021trust} and the \textit{sequential policy update scheme}~\cite{kuba2021trust} to implement multi-agent trust-region learning with monotonic improvement guarantee.
    \item \textbf{HAPPO} is the first-order emulation algorithm of HATRPO that follows the idea of PPO.
\end{itemize}

We further compare the proposed optimistic MAPG method with existing optimistic Q-learning based methods. In \textit{Multi-agent MuJoCo} with continuous action space, we include the recent \textbf{FACMAC}~\cite{peng2021facmac} as our optimistic baseline which overcomes the \textit{RO} problem by using the other agents' newest policy to update individual policy. \textbf{Hysteretic DQN}~\cite{matignon2007hysteretic, omidshafiei2017deep} is used as our optimistic baseline in the \textit{Overcooked} domain which has discrete action space.

The RO problem can be seen as a special category of the general exploration problem. Therefore, we also investigate whether existing multi-agent exploration methods can solve the RO problem. \textbf{NA-MAPPO}~\cite{hu2021policy} is a general exploration method by injecting noise into the advantage estimates. \textbf{MAVEN}~\cite{mahajan2019maven} facilitates coordinated exploration by conditioning the joint Q-value on a latent state. \textbf{CMAE}~\cite{liu2021cooperative} learns a separate exploration policy based on the count of the visitations of states.

Our method is implemented based on MAPPO and we use the same hyperparameters as MAPPO in all tasks. For HAPPO and HATRPO, we follow their original implementation and hypermeters but align the learning rate and the number of rollout threads to have a fair comparison. The implementation details including the pseudo-code of OptiMAPPO can be found in Appendix~\ref{sec:details}.

\subsection{Repeated Matrix Games}
    Even though with small state and action spaces, the \textit{climbing} and \textit{penalty} matrix games~\cite{claus1998dynamics}, as shown in Figure~\ref{fig:matrix}, are usually hard to obtain the optimal joint solution without explicitly overcoming the \textit{relative overgeneralization} problem. The matrix games have two agents which select the column and row index of the matrix respectively. The goal is to select the correct row and column index to obtain the maximal element of a matrix.
    In the \textit{penalty} game, $k\leq 0$ is the penalty term and we evaluate for $k \in \{-100, -75, -50, -25, 0\}$. The lower value of $k$, the harder for agents to identify the optimal policy due to the growing risk of penalty. Following~\cite{papoudakis2020benchmarking}, we set constant observation and the episode length of repeated games as $25$.

\begingroup
\setlength{\tabcolsep}{4.8pt} % Default value: 6pt

    \begin{table}[ht]
        \centering
        \caption{The average returns of the repeated matrix games.}
        \begin{tabular}[t]{lcccc}
        \hline
        task\textbackslash algo. & MAPPO &HAPPO & HATRPO & Ours\\
        \hline
        Climbing & $175$ & $175$ & $150$ & $\mathbf{275}$\\ 
        Penalty k=0  & $\mathbf{250}$ & $\mathbf{250}$ & $\mathbf{250}$ & $\mathbf{250}$\\
        Penalty k=-25  & $50$ & $50$ & $50$ & $\mathbf{250}$\\ 
        Penalty k=-50  & $50$ & $50$ & $50$ & $\mathbf{250}$\\ 
        Penalty k=-75  & $50$ & $50$ & $50$ & $\mathbf{250}$\\ 
        Penalty k=-100  & $50$ & $50$ & $50$ & $\mathbf{250}$\\  
        
        \hline
        \end{tabular}
        \label{tab:tab2}
    \end{table}%

\endgroup

    To better understand the RO problem and the role of being optimistic in these cooperative tasks, Figure~\ref{fig:matrix} compares the learning process with and without the optimistic update on the Climbing task. Initially, both agents uniformly assign probability on each index. In each step $t$, the agents update their individual policy distribution based on $ \pi_{t+1}(i) = \textit{softmax} (\frac{Q_i}{\eta})$~\cite{abdolmaleki2018maximum,peters2010relative}. $Q_i$ is calculated as $\sum_j \pi_t(i, j) R(i, j)$, $\eta$ is fixed as 2 and $R(i, j)$ is the payoff at row $i$, column $j$. From Figure~\ref{fig:matrix}, it is clear to see after the first update step, without handling the RO issue (the first row), the joint policy quickly assigns a high probability on a sub-optimal solution and assigns low probabilities on the rest, while optimistic update avoids it. This is important in cooperative tasks to prevent premature convergence. As we can see after 10 update steps, the baseline method converges to a sub-optimal solution by selecting the number $7$ while the optimistic update successfully finds the global solution.

    In Table~\ref{tab:tab2}, we compare our method with other MAPG baselines. We can see that OptiMAPPO with optimistic update achieves the global optima, while the baseline without considering the \textit{relative overgeneralization} problem converge to the local optima. The performance of more popular MARL methods on matrix games can be found in Appendix~\ref{sec:matrix}.

\subsection{Multi-agent MuJoCo (\textit{MA-MuJoCo})}

\begin{figure*}[ht]
% \vspace{-1em}
\centering
\includegraphics[width=0.96\textwidth, ]{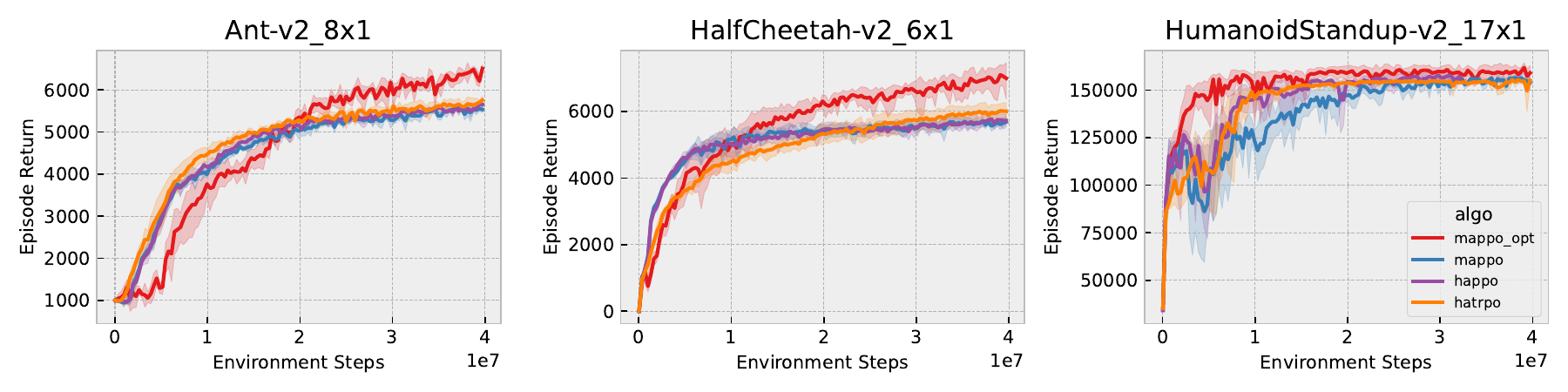}

\vspace{-1em}
\caption{Comparisons of average episodic returns on three \textit{MA-MuJoCo} tasks. OptiMAPPO converges to a better joint policy in these tasks. We plot the mean across 5 random seeds, and the shaded areas denote 95\% confidence intervals.}

\label{fig: mujoco_part}
\end{figure*}

In this section, we investigate whether the proposed OptiMAPPO can scale to more complex continuous tasks and how it compares with the state-of-the-art MAPG methods. \textit{MA-MuJoCo}~\cite{peng2021facmac} contains a set of complex continuous control tasks which are controlled by multiple agents jointly. The evaluation results of the selected three tasks are presented in Figure~\ref{fig: mujoco_part} and the full results on all the $11$ tasks are left in the Appendix~\ref{sec:mujoco_full}.
We observe that OptiMAPPO obtains clearly better asymptotic performance in most tasks compared to the baselines. 
% Particularly, in the \textit{Ant} and \textit{HalfCheetah} tasks, as the number of agents increases, OptiMAPPO outperforms the baselines more. 

In the \textit{Humannoid Standup} task which needs $17$ agents to coordinate well to stand up, the baselines experience drastic oscillation during learning while OptiMAPPO increases much more stably. We would like to attribute the oscillation in the baselines to the RO problem as the strong simultaneous coordination between agents is necessary to stand up. We also compare the maximum episodic returns during evaluation in
% Figure~\ref{fig: mujoco_max} and Figure~\ref{fig: human_max} in 
Appendix, where our method outperforms the baselines by a large margin on most tasks.

We observe that at the beginning of the training, our method learns more slowly than the baselines on the \textit{MA-MuJoCo} tasks, however, this is not as evident on the \textit{Overcooked} tasks. This is because the advantage clipping $\eta=0$ disregards the information of the data with negative advantages, which can be seen as a cost to converge to a better solution. However, on \textit{Overcooked}, since the task spaces are smaller compared to \textit{MA-MuJoCo} tasks, the samples are sufficient in each update.
% for optimistic updating. 
\begin{figure*}[ht]
\centering
\includegraphics[width=0.96\textwidth, ]{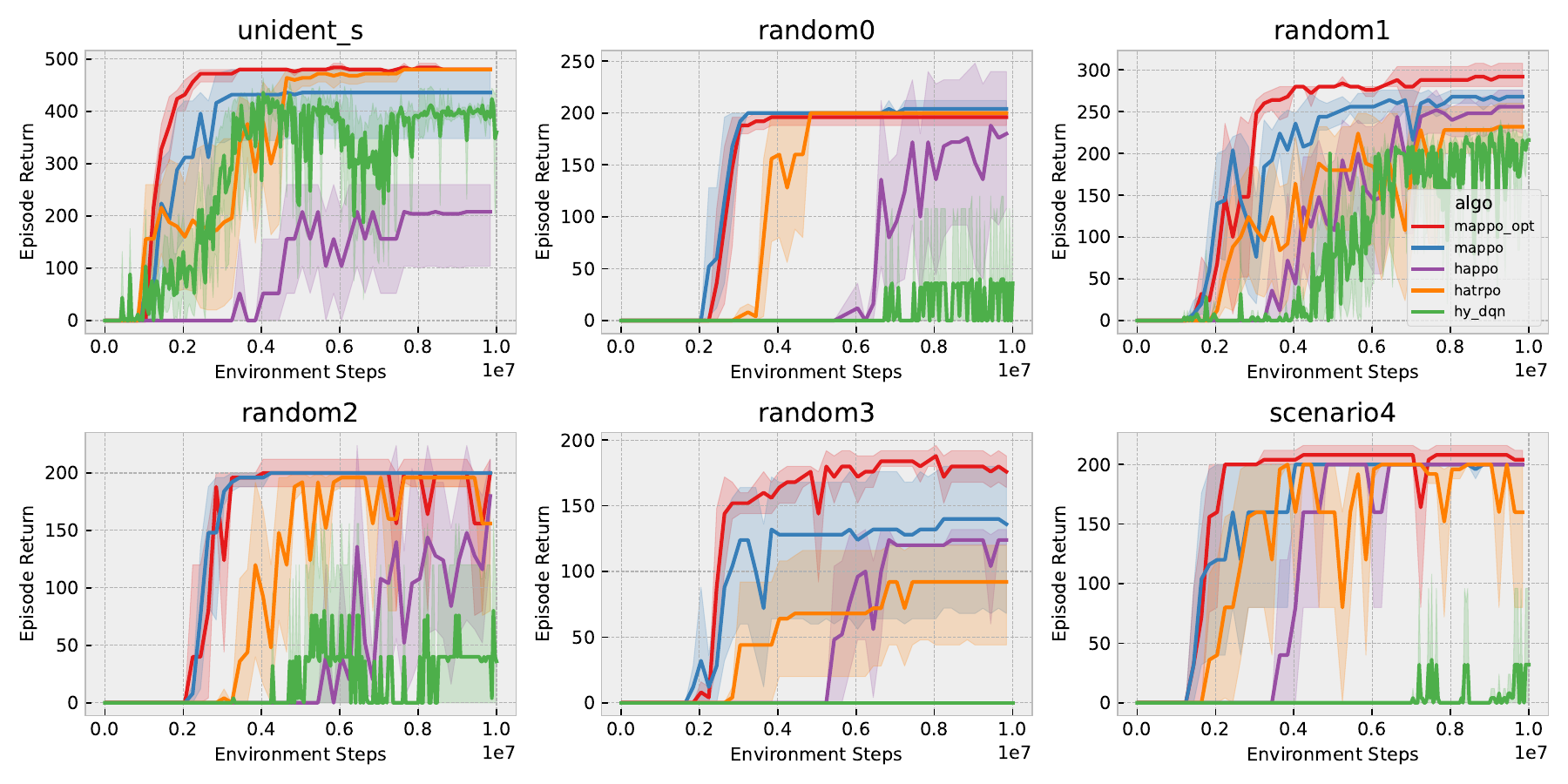}

\vspace{-1.2em}
\caption{Comparisons of average episodic returns on \textit{Overcooked} tasks. Our method outperforms or matches strong baselines and hysteretic DQN (\textit{hy\_dqn} in the legend) on tested tasks. Although with optimism, \textit{hy\_dqn} fails to boost good performance. 
% We plot the mean across 5 random seeds, and the shaded areas denote 95\% confidence intervals.
}

\label{fig: overcooked}
\end{figure*}

\begingroup
\setlength{\tabcolsep}{6pt} % Default value: 6pt

\begin{table}[ht]
\caption{The comparison with FACMAC on nine of \textit{MA-MuJoCo} tasks. We list the average episode return and the standard deviation. The bold number indicates the best. \textit{HalfCh} is short for HalfCheetah.}
% \vskip -0.15in
\[
\begin{array}{@{}l*{2}{l}@{}}
\toprule
\text{Task} & \multicolumn{2}{c@{}}{\text{Algorithm}}\\
    \cmidrule(l){2-3}
    & \text{FACMAC}\;(\sigma) & \text{OptiMAPPO}\;(\sigma) \\
\midrule
\text{Ant 2x4} & 307.58\; (78.28) & \mathbf{6103.97\; (180.62)}\\
\text{Ant 4x2} & 1922.26\; (285.94) & \mathbf{6307.75\; (114.74)} \\
\text{Ant 8x1} & 1953.04\; (2276.16) & \mathbf{6393.07\; (59.11)}\\
\text{Walker 2x3} & 713.34\; (600.01) & \mathbf{4571.36\; (262.40)}\\
\text{Walker 3x2} & 1082.23\; (572.40) & \mathbf{4582.90\;(143.01)}\\
\text{Walker 6x1} & 950.05\; (542.33) & \mathbf{4957.02 \; (650.93)}\\
\textit{HalfCh 2x3} & 5069.17\; (2791.02) & \mathbf{6499.82\;(573.55)}\\
\textit{HalfCh 3x2} & 5379.35\; (4229.25) & \mathbf{6887.77 \;(406.89)}\\
\textit{HalfCh 6x1} & 3482.91\; (3374.16) & \mathbf{6982.65\;(490.35)}\\
\bottomrule
\end{array}
\]
\vskip -0.15in
\label{tab:facmac}
\end{table}

\endgroup
    
\paragraph{Comparison with FACMAC} We include FACMAC as our optimistic baseline~\cite{peng2021facmac} for the continuous action space tasks. FACMAC has been proposed to improve MADDPG~\cite{lowe2017multi} by solving the \textit{RO} problem with a centralized gradient estimator. Specifically, FACMAC samples all actions from agents’ current policies when evaluating the joint action-value function, which can be seen as one way to achieve optimism since the newest policy would generate higher Q estimation. As the results listed in Table~\ref{tab:facmac}, OptiMAPPO significantly outperforms FACMAC, which shows that our method can better employ the advantage of optimism and achieve stronger performance than the current optimistic method.

\subsection{Overcooked}

\textit{Overcooked}~\cite{carroll2019utility, yu2023learning} is a fully observable two-player cooperative game that requires the agents to coordinate their task assignment to accomplish the recipe as soon as possible. We test OptiMAPPO on $6$ tasks with different layouts.
To succeed in these games, players must coordinate to travel around the kitchen and alternate different tasks, such as collecting onions, depositing them into cooking pots, and collecting a plate.

\begin{figure*}[ht]
\centering
\includegraphics[width=0.32\textwidth, ]{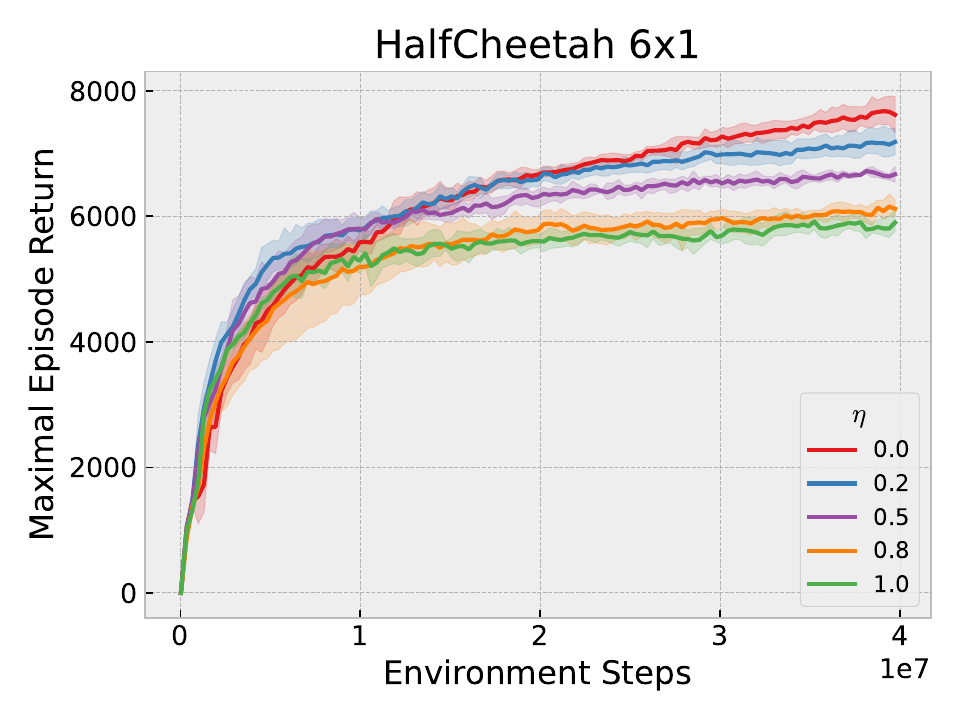}
\includegraphics[width=0.32\textwidth, ]{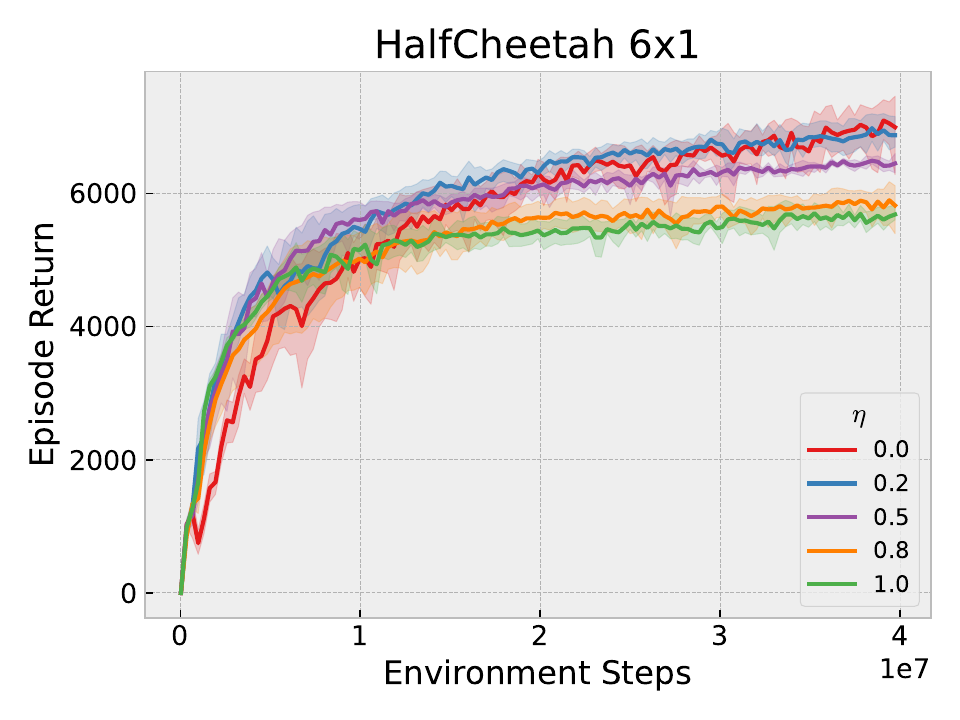}
\includegraphics[width=0.32\textwidth, ]{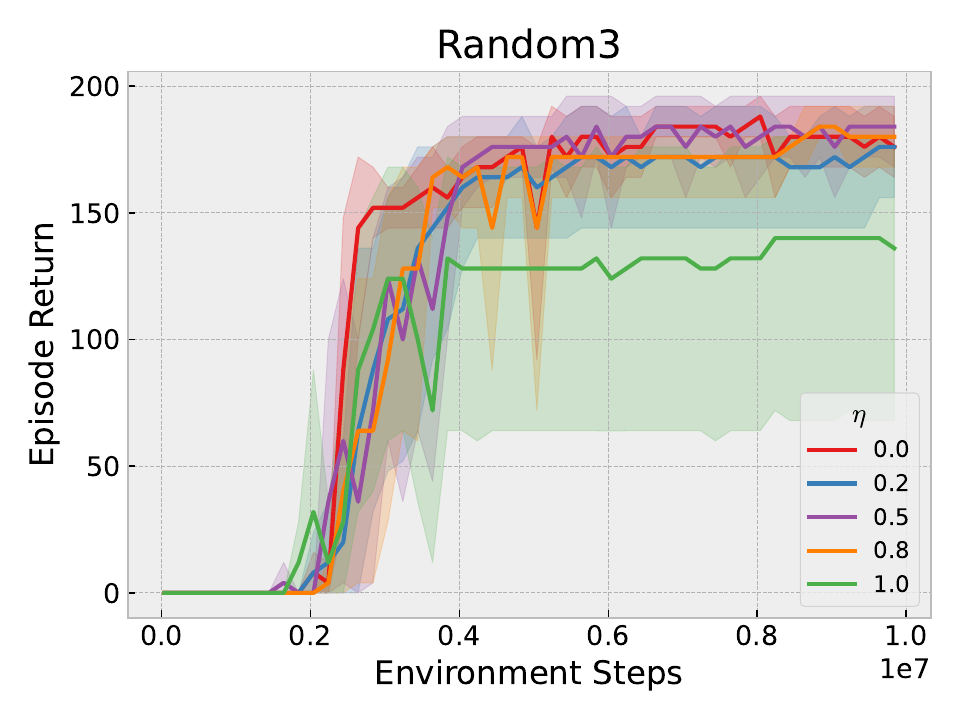}
% \vspace{-1em}
\caption{Ablation experiments on different degrees of optimism in OptiMAPPO. It shows that optimism helps in both tasks to a wide range of degrees. Particularly, in \textit{HalfCheetah 6x1}, with decreasing $\eta$, i.e. increasing degree of optimism, the performance gradually improves.}
% \vskip -0.2in
\label{fig: ablation}
\end{figure*}

The comparisons with both recent MAPG algorithms and existing optimism baseline are shown in Figure~\ref{fig: overcooked}, where our method consistently achieves similar or better performance on all tasks. For example in the \textit{Random3} task, We observe that the baseline methods either consistently converge to a suboptimal policy (in HAPPO), or show a big variance between different training (in MAPPO). In the rendered videos, the suboptimal policy in HAPPO can only learn to assign one agent to deliver, while our method successfully learns to rotate in a circle to speed up the delivery. We speculate that the gap between our method and baselines is due to the presence of subtle RO problems in certain tasks. The baseline methods can only rely on naive exploration to find the optimal joint policy, which is susceptible to the RO problem. However, our method with advantage shaping can effectively overcome these issues and converge more stably. In Section~\ref{sec:hyq}, even though the optimistic baseline hysteretic DQN also utilizes optimism to overcome the \textit{RO} problem, it can incur drastic overestimation and thus fails to improve the performance, as shown in our experiments.

\subsection{Comparison with Exploration Methods}
The results of MAVEN~\cite{mahajan2019maven} and NA-MAPPO~\cite{hu2021policy} on the matrix games are shown in Table~\ref{tab:exploration}, where both methods fail to solve the RO problem. The experiments show that general exploration methods following the principle of being optimistic in the face of uncertainty~\cite{munos2014bandits, imagawa2019optimistic} may not be able to solve the RO problem. Our method works by being optimistic to the suboptimal joint actions instead of unseen states or actions. Comparisons with NA-MAPPO in the \textit{MA-MuJoCo} domain can be found in Appendix~\ref{sec:namppo}.

\begin{table}[ht]
        \centering
        \caption{Performance of General Exploration Methods.}
        \begin{tabular}[t]{lccc}
        \hline
        task\textbackslash algo. & MAVEN &NA-MAPPO & Ours\\
        \hline
        Climbing & $175$ & $175$ & $\mathbf{275}$\\ 
        Penalty k=0  & $\mathbf{250}$ & $\mathbf{250}$ & $\mathbf{250}$\\
        Penalty k=-25  & $50$ & $50$ & $\mathbf{250}$\\ 
        Penalty k=-50  & $50$ & $50$ & $\mathbf{250}$\\ 
        Penalty k=-75  & $50$ & $50$ & $\mathbf{250}$\\ 
        Penalty k=-100  & $50$ & $50$ & $\mathbf{250}$\\  
        
        \hline
        \end{tabular}
        \label{tab:exploration}
    \end{table}%
    
We also compare our algorithm with CMAE~\cite{liu2021cooperative} on \textit{Penalized Push-Box}. This benchmark modifies the original \textit{Push-Box} task in~\cite{liu2021cooperative} by injecting penalty to agents when the agents are not coordinated to push box at the same time, thereby presenting the RO issue. Table~\ref{tab:pushbox} shows that both our method and CMAE can successfully overcome the RO problem and converge to the global optima (episode return as $1.6$) while the vanilla MAPPO fails. However, note that CMAE is implemented on a tabular Q representation and suffers from exponentially increasing complexity, while our method can scale up well.
    \begin{table}[ht]
        \centering
        \caption{Results on \textit{Penalized Push-Box}.}
        \begin{tabular}[t]{lccc}
        \hline
        task\textbackslash algo. & MAPPO &CMAE & Ours\\
        \hline
        \textit{Penalized Push-Box} & $0$ & $1.6$ & $1.6$\\ 
        
        \hline
        \end{tabular}
        \label{tab:pushbox}
    \end{table}%

\subsection{How Much Optimism Do We Need?}

In this section, we perform an ablation study to examine how the performance change when we gradually change the degree of optimism, i.e. setting different values of $\eta$ in the \textit{Leaky ReLU} extension. We experiment $\eta=\{0.2, 0.5, 0.8\}$ on the \textit{HalfCheetah 6x1} in \textit{MA-MuJoCo} and \textit{Random 3} in \textit{Overcooked}. Note that OptiMAPPO takes $\eta=0$ and degrades to MAPPO when $\eta=1$. The results shown in Figure~\ref{fig: ablation} indicate that optimism helps in both tasks to a wide range of degrees. While in the \textit{HalfCheetah 6x1} task, the performance improvement turns out clearly proportional to the degree of optimism.
In \textit{Random 3} task, even when $\eta=0.8$, OptiMAPPO converges as well as the full optimism. This can be partially due to the rewards in \textit{Overcooked} being discretized in a coarse grain.
Overall, the experiments augment our hypothesis that the optimism helps to overcome the \textit{relative overgeneralization} in multi-agent learning and eventually helps to converge to a better joint policy.

\subsection{Does Q-learning Based Optimism Work?}\label{sec:hyq}

\begin{figure}[ht]
% \vskip -0.15in
\centering
    \includegraphics[width=0.48\textwidth]{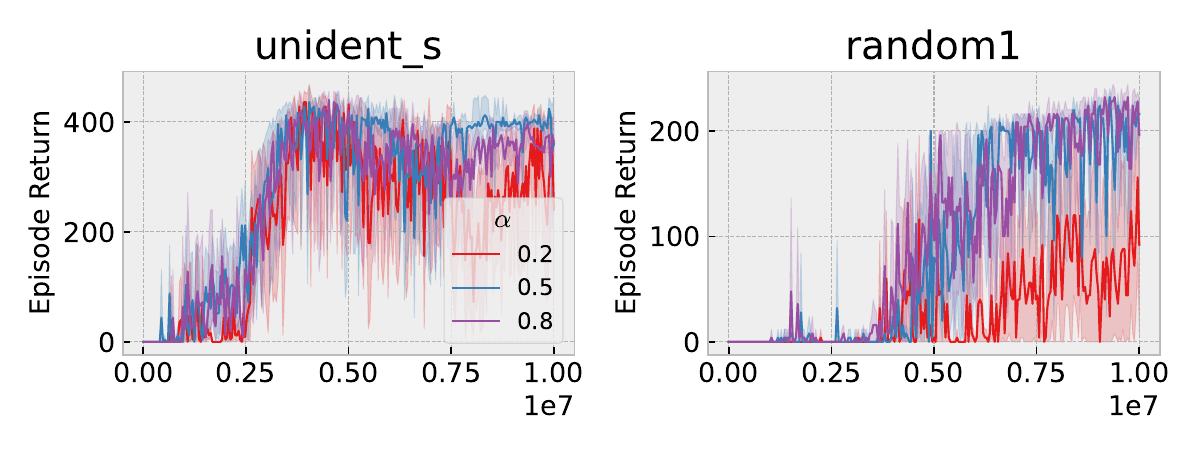}
    % \vskip -0.17in
    \includegraphics[width=0.48\textwidth]{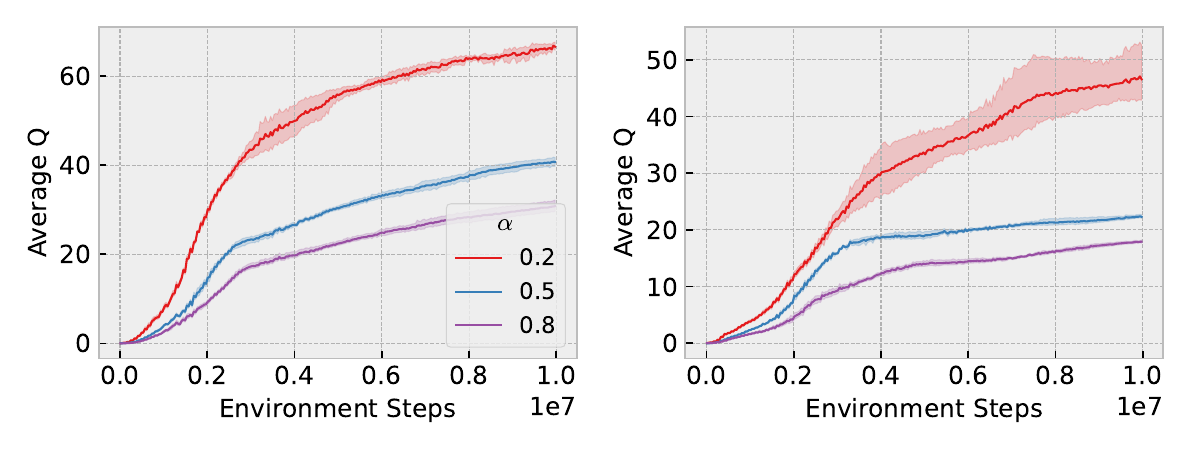}

% \includegraphics[width=0.48\textwidth, ]{figs/overcooked_lambda_1.pdf}
% \vspace{-1em}
\caption{The up row shows the episode return of hysteretic DQN with different $\alpha$, while the corresponding average Q values are shown in the bottom row. The Q values gradually increase with increasing degree of optimism, i.e. lower $\alpha$, which may degrade the performance.
}
% \vskip -0.15in
\label{fig: overcooked_lambda}
\end{figure}

 We empirically analyze whether Q-learning based optimism can boost performance in the \textit{Overcooked} tasks using hysteretic Q-learning~\cite{omidshafiei2017deep, palmer2017lenient}. Figure~\ref{fig: overcooked_lambda} shows average Q values during training with different degrees of optimism. When $\alpha$ decreases excessively, i.e.\  optimism increases too much, performance decreases. In both tasks shown in Figure~\ref{fig: overcooked_lambda}, the highest optimism estimates the highest Q-values while showing the worst performance.

Different from the Q-learning based optimism methods, our proposed optimistic MAPG method performs the advantage estimation in an on-policy way and thus circumvents the overestimation problem naturally. Therefore, our method can fully employ the advantage of optimism and demonstrates strong performance in complex domains.

\section{Limitation and Conclusion}
Our optimistic updating approach yields state-of-the-art performance. However, as with other optimistic updating methods~\cite{lauer2000algorithm, matignon2007hysteretic}, optimism can lead to a sub-optima when misleading stochastic rewards exist. The proposed \textit{Leaky ReLU} extension allows further adaptive adjustment of the optimism degree $\eta$ to balance between optimism and neutrality and may allow to reduce the severity of stochastic rewards but this requires future investigation. In addition, lenient agents~\cite{panait2006lenient, wei2016lenient, palmer2017lenient} provide a set of heuristic techniques to adapt the degree of optimism, applicable to our method, to mitigate the problem with stochastic rewards. We leave this also as future work.

Motivated by solving the \textit{relative overgeneralization} problem, we investigate the potential of optimistic updating in state-of-the-art MAPG methods. We first introduce a general advantage reshaping approach to incorporate optimism in policy updating, which is easy to implement based on existing MAPG methods. To understand the proposed advantage transformation, we provide a formal analysis from the operator view of policy gradient methods. The analysis shows that the proposed advantage shaping retains the optimality of the policy at a fixed point. Third, we extensively evaluate the instantiated optimistic updating policy gradient method, OptiMAPPO. Experiments on a wide variety of complex benchmarks show improved performance compared to state-of-the-art baselines with a clear margin. 

\section*{Acknowledgements}

% \textbf{Do not} include acknowledgements in the initial version of
% the paper submitted for blind review.

% If a paper is accepted, the final camera-ready version can (and
% usually should) include acknowledgements.  Such acknowledgements
% should be placed at the end of the section, in an unnumbered section
% that does not count towards the paper page limit. Typically, this will 
% include thanks to reviewers who gave useful comments, to colleagues 
% who contributed to the ideas, and to funding agencies and corporate 
% sponsors that provided financial support.
We acknowledge the computational resources provided by
the Aalto Science-IT project and CSC, Finnish IT Center
for Science, and, funding by Research Council of Finland (353138, 327911, 357301). We also thank the ICML reviewers for the suggestions to connect our work to coordinated exploration methods.

\section*{Impact Statement}
This paper presents work whose goal is to advance the field of 
Machine Learning. There are many potential societal consequences 
of our work, none which we feel must be specifically highlighted here.

% Authors are \textbf{required} to include a statement of the potential 
% broader impact of their work, including its ethical aspects and future 
% societal consequences. This statement should be in an unnumbered 
% section at the end of the paper (co-located with Acknowledgements -- 
% the two may appear in either order, but both must be before References), 
% and does not count toward the paper page limit. In many cases, where 
% the ethical impacts and expected societal implications are those that 
% are well established when advancing the field of Machine Learning, 
% substantial discussion is not required, and a simple statement such 
% as the following will suffice:

% ``This paper presents work whose goal is to advance the field of 
% Machine Learning. There are many potential societal consequences 
% of our work, none which we feel must be specifically highlighted here.''

% The above statement can be used verbatim in such cases, but we 
% encourage authors to think about whether there is content which does 
% warrant further discussion, as this statement will be apparent if the 
% paper is later flagged for ethics review.


\bibliography{main}
\bibliographystyle{icml2024}

%%%%%%%%%%%%%%%%%%%%%%%%%%%%%%%%%%%%%%%%%%%%%%%%%%%%%%%%%%%%%%%%%%%%%%%%%%%%%%%
%%%%%%%%%%%%%%%%%%%%%%%%%%%%%%%%%%%%%%%%%%%%%%%%%%%%%%%%%%%%%%%%%%%%%%%%%%%%%%%
% APPENDIX
%%%%%%%%%%%%%%%%%%%%%%%%%%%%%%%%%%%%%%%%%%%%%%%%%%%%%%%%%%%%%%%%%%%%%%%%%%%%%%%
%%%%%%%%%%%%%%%%%%%%%%%%%%%%%%%%%%%%%%%%%%%%%%%%%%%%%%%%%%%%%%%%%%%%%%%%%%%%%%%
\newpage
\appendix
\onecolumn

% You can have as much text here as you want. The main body must be at most $8$ pages long.
% For the final version, one more page can be added.
% If you want, you can use an appendix like this one.  

% The $\mathtt{\backslash onecolumn}$ command above can be kept in place if you prefer a one-column appendix, or can be removed if you prefer a two-column appendix.  Apart from this possible change, the style (font size, spacing, margins, page numbering, etc.) should be kept the same as the main body.

% \section{Appendix}
\section{Operator View of Policy Gradient}\label{sec:operator_intro}
As shown in~\cite{ghosh2020operator}, the policy update in vanilla policy gradient can be seen as doing a gradient step to minimize

% The proposed advantage shaping aims to enable optimism in MAPG methods. This transformation might be hard to connect to existing approaches. Therefore, we fit the advantage shaping into a range of popular methods from an operator view of policy gradient following the derivations in~\cite{ghosh2020operator}. As shown in~\cite{ghosh2020operator}, the policy update in policy gradient can be seen as doing a gradient step to minimize

\begin{equation}\label{equ: gradient_step}
    D_{V^{\pi_t}\pi_t}(Q^{\pi_t}\pi_t \lvert\rvert \pi)=\sum_s d^{\pi_t}(s)V^{\pi_t}(s)  \text{KL}(Q^{\pi_t}\pi_t\lvert\rvert \pi),
\end{equation}
where $d^{\pi}(s)$ is the discounted stationary distribution induced by the policy $\pi$. $D_{V^{\pi_t}\pi_t}$ and the distribution $Q^{\pi_t}\pi$ over actions are defined as
\begin{equation}\label{equ: min_two_steps}
\begin{aligned}
    &D_{z}(\mu\lvert\rvert \pi)=\sum_s z(s)\text{KL}(\mu(\cdot\vert s)\lvert\rvert \pi(\cdot \vert s)), \\
    &Q^{\pi}\pi(a\vert s)=\frac{1}{V^{\pi}(s)}Q^{\pi}(s, a)\pi(a\vert s),
\end{aligned}
\end{equation}
% \begin{equation}
%     Q^{\pi}\pi(a\vert s)=\frac{1}{V^{\pi}(s)}Q^{\pi}(s, a)\pi(a\vert s),
% \end{equation}
which corresponds to two successive operation by the \textit{projection operator} $\mathcal{P}_V$ and the \textit{improvement operator} $\mathcal{I}_V$,
\begin{equation}\label{equ: operators}
\begin{aligned}
    &\mathcal{I}_V\pi(s,a)=(\frac{1}{\mathbb{E}_{\pi}[V^{\pi}]}d^{\pi}(s)V^{\pi}(s))Q^{\pi}\pi(a\vert s), \\ 
    &\mathcal{P}_{V\mu}=\argmin_{z\in \Pi}\sum_s\mu(s)\text{KL}(\mu(\cdot \vert s)\lvert\rvert z(\cdot \vert s)).
\end{aligned}
% \vskip -0.2in
% \label{equ: improve}
% \vskip -0.2in
\end{equation}

% \begin{equation}\label{equ:projection}
%     \mathcal{P}_{V\mu}=\argmin_{z\in \Pi}\sum_s\mu(s)\text{KL}(\mu(\cdot \vert s)\lvert\rvert z(\cdot \vert s)),
% \end{equation}

The \textit{improvement operator} $\mathcal{I}_V$ tries to improve the policy into general function space $\mu(\cdot \vert s)$ via the information provided by the Q values, while the \textit{projection operator} $\mathcal{P}_V$ projects the $\mu(\cdot \vert s)$ into the policy function space $\pi(a\vert s)$. In this way, the policy gradient and Q-learning can be connected using the same polynomial operator $\mathcal{I}_V = (Q^{\pi})^{\frac{1}{\alpha}}\pi$, where the REINFORCE~\cite{williams1992simple} is recovered by setting $\alpha=1$ and Q-learning is obtained at the limit $\alpha=0$. 

More sophisticated policy gradient methods arise by designing different $\mathcal{I}_V$ which constructs different candidate distributions before being projected to the policy function space. For example, MPO~\cite{abdolmaleki2018maximum} uses a normalized exponential of Q values $\exp(\beta Q^{\pi}(s,a))$. 

\section{Proof of Proposition~\ref{thm:thm2}}\label{sec:proof}

\begin{proof}[\textbf{Proof of Proposition}~\ref{thm:thm2}]\label{proof:operator}
Following the proof for the regular policy gradient in ~\cite{ghosh2020operator}, we replace the non-negative Q function with clipped advantages, which guarantees a valid probability distribution in the projection operators. Therefore, we have

\begin{equation}
    \begin{aligned}
    &\qquad\nabla_{\theta}\sum_s d^{\pi^{\ast}(s)}V^{+\pi^{\ast}}(s)\text{KL}(\text{clip}(A^{\pi^{\ast}})\pi^{\ast}\lvert\rvert \pi)\mid_{\pi=\pi^{\ast}}\\
    &=\sum_s d^{\pi^{\ast}}(s)\sum_a \pi^{\ast}(a\vert s)\text{clip}(A^{\pi^{\ast}}(s,a))\frac{\partial \log \pi_{\theta}(a\vert s)}{\partial \theta}\mid_{\theta=\theta^{\ast}}\\
    &=0 \text{ by definition of }\pi^{\ast},
\end{aligned}
\end{equation} 
where $\text{clip}(A^{\pi})\pi$ and $V^{+\pi}$ are defined as:
\begin{equation}
\begin{aligned}
    &\text{clip}(A^{\pi})\pi(a\vert s)=\frac{1}{V^{+\pi}(s)}\text{clip}(A^{\pi}(s,a))\pi(a\vert s),\\ &V^{+\pi}(s)=\sum_a \text{clip}(A(s,a))\pi(a\vert s).
\end{aligned}
\end{equation}

% \begin{align}
%     \nabla_{\theta}\sum_s d^{\pi^{\ast}(s)}V^{\pi^{\ast}}(s)\text{KL}(\text{LR}(A^{\pi^{\ast}})\pi^{\ast}\lvert\rvert \pi)\mid_{\pi=\pi^{\ast}}&=\sum_s d^{\pi^{\ast}}(s)\sum_a \pi^{\ast}(a\vert s)LR(A^{\pi^{\ast}}(s,a))\frac{\partial \log \pi_{\theta}(a\vert s)}{\partial \theta}\mid_{\theta=\theta^{\ast}}\\
%     &=0.
% \end{align}
\end{proof}

\section{Implementation Details}\label{sec:details}
We introduce the important implementation details here and the full details can be found in our code.

\subsection{Pseudo Code for OptiMAPPO}\label{sec:pseudo_code}

\begin{algorithm}[H]
   \caption{Optimistic Multi-Agent Proximal Policy Optimization (OptiMAPPO)}
   \label{alg:spmarl}
\begin{algorithmic}
   \STATE {\bfseries Input:} Initialize value function $V_{\phi}(s)$, individual policies ${\pi_{\theta^i}(a^i\vert s^i)}, i\in \{1,\cdots, n\}$, buffer $\mathcal{D}$, iteration K, samples per iteration M
   \FOR{$k=1$ {\bfseries to} $K$}
%   \IF{$x_i > x_{i+1}$}
    \STATE \textbf{Collect on-policy data:}
    \FOR{$i=1$ {\bfseries to} $M$}
    \STATE Sample individual actions $\{a_1, \cdots, a_n\}$ from individual policies $\{\pi_{1}, \cdots, \pi_{n}\}$ 
    \STATE Interact with the environment and collect trajectories $\tau$ into buffer $\mathcal{D}$
    \ENDFOR
    \STATE \textbf{Policy update:}
    \STATE Estimate state values $V(s)$ and advantages $A(s,a_i)$ as Equation~\ref{equ:gae} 
    \STATE Clip the negative advantages to get $\text{clip}(A(s,a_i), 0)$
    \STATE Update value function $V_{\phi}(s)$ by minimizing Equation~\ref{equ:td}
    \STATE Optimize policies following Equation~\ref{equ:policy_update}
   
%   \ENDIF
   \ENDFOR
   
\end{algorithmic}
\end{algorithm}

\subsection{Key Hyper-Parameters}

\paragraph{Repeated Matrix Games:}
In the two repeated matrix games, since the observation for each time step is fixed as a constant and the learned state value function is uninformative, we compute the advantage using TD(0) in both OptiMAPPO and MAPPO, instead of the GAE in order to reduce the noise.

\paragraph{\textit{MA-MuJoCo}:}
In all the tasks of \textit{MA-MuJoCo}, we use the same hyperparameters listed in Table~\ref{tab:hyper_mujoco}. The implementation is based on the HAPPO~\cite{kuba2021trust} codebase, and the other hyperparameters are the default.

\begin{table}[ht]
\caption{Key Hyper-parameters for the \textit{MA-MuJoCo} tasks.}
\label{tab:hyper_mujoco}
% \vskip 0.15in
\begin{center}
% \begin{small}
% \begin{sc}
\begin{tabular}{lccccr}
\toprule
Hyper-parameters & MAPPO & OptiMAPPO & HAPPO & HATRPO \\
\midrule
Recurrent Policy    & No  & No & No & No\\
Parameter Sharing & No & No & No & No\\
Episode Length    &$1000$ &$1000$ &$1000$ &$1000$ \\
No. of Rollout Threads    & $32$ & $32$ & $32$ & $32$\\
No. of Minibatch    & $40$ & $40$ & $40$ & $40$ \\
Policy Learning Rate   & $0.00005$ & $0.00005$ & $0.00005$ & $0.00005$\\
Critic Learning Rate   & $0.005$ & $0.005$ & $0.005$ & $0.005$\\
Negative Slope $\eta$  & N/A & $0$ & N/A  & N/A\\
KL Threshold  & N/A & N/A & N/A &$0.0001$ \\
\bottomrule
\end{tabular}
% \end{sc}
% \end{small}
\end{center}
% \vskip -0.1in
\end{table}

\paragraph{\textit{Overcooked}:} In all the tasks of \textit{Overcooked}, we use the same hyperparameters listed in Table~\ref{tab:hyper_overcooked}.  

\begin{figure}[ht]
    \centering
    \subfigure[\textit{Random3}]{
        \includegraphics[width=0.24\textwidth]{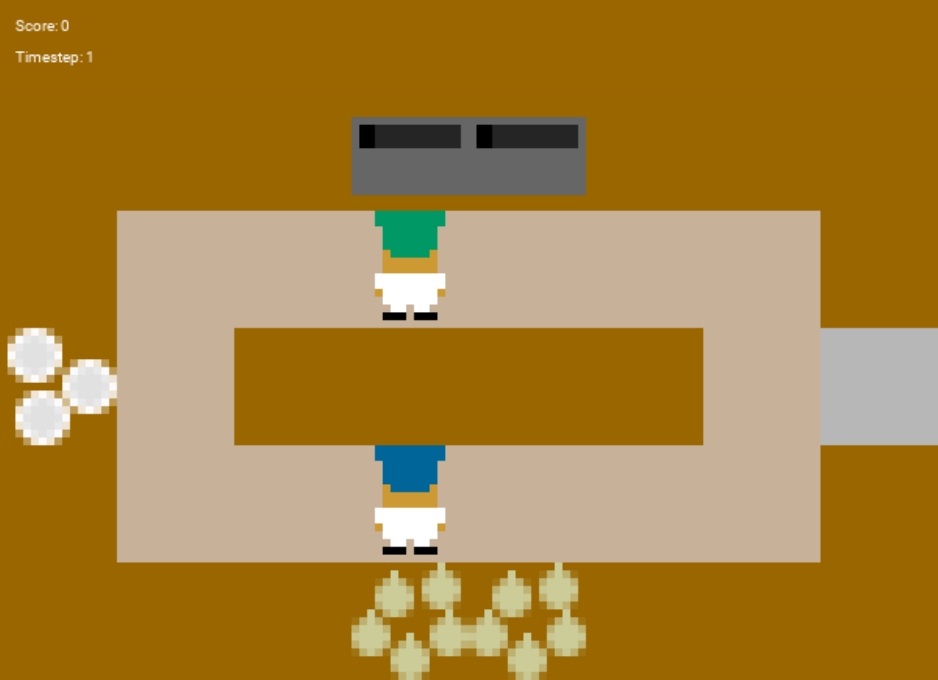}
        \label{fig:Random3}
    }
    \subfigure[\textit{Unident\_s}]{
        \includegraphics[width=0.24\textwidth]{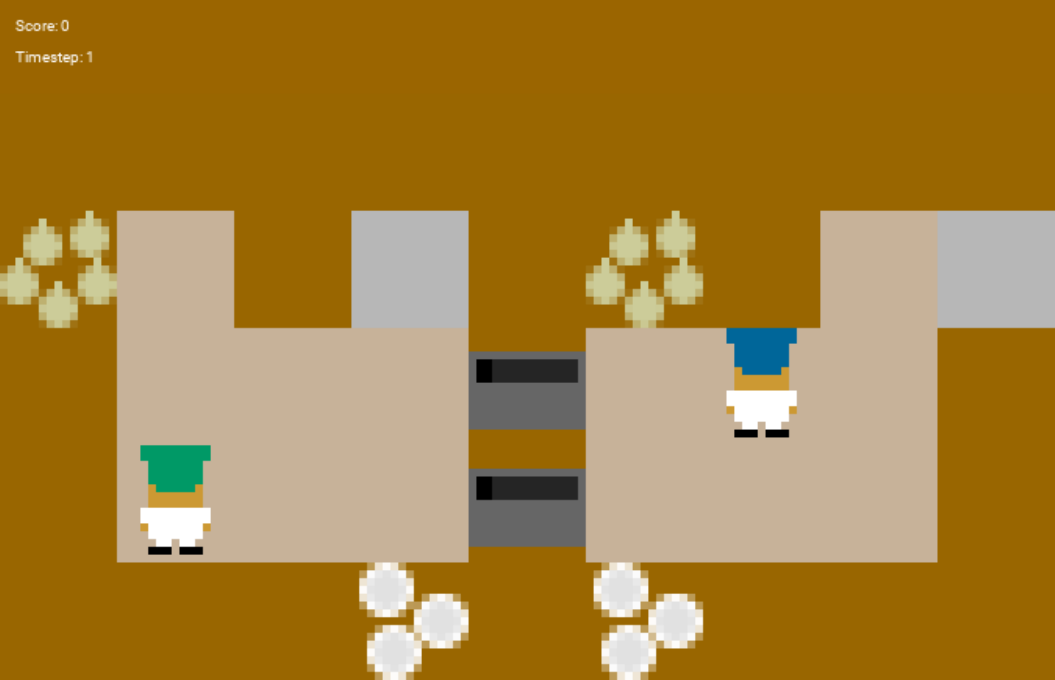}
        \label{fig:Unident}
    }
    \caption{The layout of two tasks in the \textit{Overcooked}.}
    \label{fig:overcooked_example}
\end{figure}

Two example tasks in \textit{Overcooked} are shown in Figure~\ref{fig:overcooked_example}. In task \textit{Random3}, the agents need to learn to circle through the corridor while avoiding blocking each other. In \textit{Unident\_s}, the agents learn to collaborate using their closest items to complete the recipe in an efficient way. However, each agent can also finish their own recipe without collaboration.

% \begin{table}[ht]
%         \centering
%         \caption{The average returns of the repeated matrix games.}
%         \begin{tabular}[t]{lcccc}
%         \hline
%         task\textbackslash algo. & MAPPO &HAPPO & HATRPO & Ours\\
%         \hline
%         Climbing & $175$ & $175$ & $150$ & $\mathbf{275}$\\ 
%         Penalty k=0  & $\mathbf{250}$ & $\mathbf{250}$ & $\mathbf{250}$ & $\mathbf{250}$\\
%         Penalty k=-25  & $50$ & $50$ & $50$ & $\mathbf{250}$\\ 
%         Penalty k=-50  & $50$ & $50$ & $50$ & $\mathbf{250}$\\ 
%         Penalty k=-75  & $50$ & $50$ & $50$ & $\mathbf{250}$\\ 
%         Penalty k=-100  & $50$ & $50$ & $50$ & $\mathbf{250}$\\  
        
%         \hline
%         \end{tabular}
%         \label{tab:tab2}
%     \end{table}%

\begin{table*}[ht]
\centering
\caption{Key Hyper-parameters for the \textit{Overcooked} tasks.}
\label{tab:hyper_overcooked}
% \vskip -0.15in
% \begin{center}
% \begin{small}
% \begin{sc}
\begin{tabular}{lcccc}
\hline
% \toprule
Hyper-parameters & MAPPO & OptiMAPPO & HAPPO & HATRPO \\
% \midrule
\hline
Recurrent Policy    & No  & No & No & No\\
Parameter Sharing & Yes & Yes & No & No\\
Episode Length    &$400$ &$400$ &$400$ &$400$ \\
No. of Rollout Threads    & $100$ & $100$ & $100$ & $100$\\
No. of Minibatch    & $2$ & $2$ & $2$ & $2$ \\
Policy Learning Rate   & $0.00005$ & $0.00005$ & $0.00005$ & $0.00005$\\
Critic Learning Rate   & $0.005$ & $0.005$ & $0.005$ & $0.005$\\
Negative Slope $\eta$  & N/A & $0$ & N/A  & N/A\\
KL Threshold  & N/A & N/A & N/A &$0.01$ \\
% \bottomrule
\hline
\end{tabular}
% \end{sc}
% \end{small}
% \end{center}
% \vskip -0.1in
\end{table*}

% \subsection{Results on More \textit{MA-MuJoCo} Tasks}

\paragraph{FACMAC:}
We use the original code base from FACMAC~\cite{peng2021facmac} with default hyper-parameter.

\paragraph{Hysteretic DQN:}
We implement Hysteretic DQN using the same state representation network as our method and select the best performance from three different $\alpha$ values: $\{0.2, 0.5, 0.8\}$ on all the \textit{Overcooked} tasks.

\section{More Results of Popular Deep MARL methods on Matrix Games}\label{sec:matrix}

In order to show the \textit{RO} problem in existing MARL methods, we cite the following results of common deep MARL methods on the \textit{penalty} and \textit{climbing} games from the benchmarking paper~\cite{papoudakis2020benchmarking}. As shown in Table~\ref{tab:marl_game}, popular MARL methods without explicitly considering the \textit{RO} problem fail to solve the matrix games. 
% \begin{table}[ht]
%         \centering
%         \caption{The average return for the repeated matrix games.}
%         \vskip 0.15in
%         \begin{tabular}[t]{lcccccccc}
%         \hline
%         task\textbackslash algo. & IQL & IA2C & MADDPG & COMA & MAA2C & MAPPO &VDN & QMIX\\
%         \hline
%         Climbing & $195$  & $175 $  & $170$ &$185$ & $175$ & $175$ & $175$ & $175$ \\ 
%         Penalty k=0  & $250$  & $250 $ & $249.98$ &$250$ & $250$ & $250$ & $250$ & $250$ \\
%         Penalty k=-25  & $50$  & $50 $  & $50$ &$50$ & $50$ & $50$ & $50$ & $50$ \\ 
%         Penalty k=-50  & $50$  & $50 $ & $50$ &$50$ & $50$ & $50$ & $50$ & $50$ \\ 
%         Penalty k=-75  & $50$  & $50 $ & $50$ &$50$ & $50$ & $50$ & $50$ & $50$ \\ 
%         Penalty k=-100  & $50$  & $50 $ & $50$ &$50$ & $50$ & $50$ & $50$ & $50$ \\  
        
%         \hline
%         \end{tabular}
%         \label{tab:marl_game}
%     \end{table}%

\begin{table*}[ht]
\caption{The average return for the repeated matrix games.}
% \vskip 0.15in
\[
\begin{array}{@{}l*{10}{c}@{}}
\hline
\text{Task} & \multicolumn{10}{c@{}}{\text{Algorithm}}\\
    \cmidrule(l){2-11}
    & \text{IQL} & \text{IA2C} & \text{MADDPG} & \text{COMA} & \text{MAA2C} & \text{MAPPO} & \text{VDN} & \text{QMIX} & \text{FACMAC} & \text{Ours} \\
\hline
\text{Climbing} & 195 & 175 & 170 & 185 & 175 & 175 & 175 & 175 & 175 & \mathbf{275} \\ 
\text{Penalty k=0} & 250 & 250 & 249.98 & 250 & 250 & 250 & 250 & 250 & 250 & \mathbf{250} \\ 
\text{Penalty k=-25} & 50 & 50 & 50 & 50 & 50 & 50 & 50 & 50 & 50 & \mathbf{250} \\
\text{Penalty k=-50} & 50 & 50 & 50 & 50 & 50 & 50 & 50 & 50 & 50 & \mathbf{250} \\
\text{Penalty k=-75} & 50 & 50 & 50 & 50 & 50 & 50 & 50 & 50 & 50 & \mathbf{250} \\
\text{Penalty k=-100} & 50 & 50 & 50 & 50 & 50 & 50 & 50 & 50 & 50 & \mathbf{250} \\
\hline
\end{array}
\]
\label{tab:marl_game}
\end{table*}

\section{Results of Optimistic MAA2C}\label{sec:maa2c}
\begin{figure}[ht]
    \centering
    \subfigure[Average episode returns]{
        \includegraphics[width=0.35\textwidth]{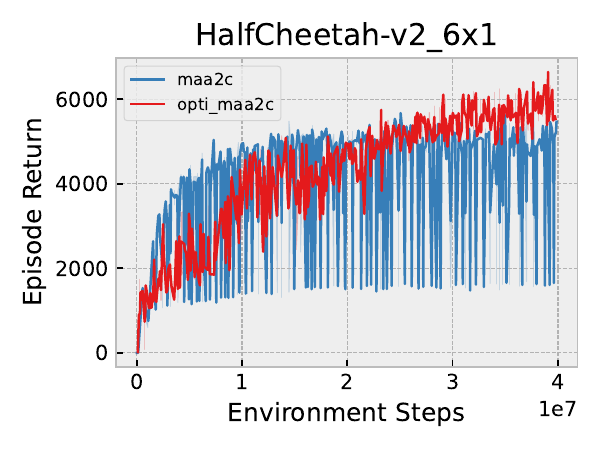}
        \label{fig:maa2c}
    }
    \subfigure[Maximum episode returns]{
        \includegraphics[width=0.35\textwidth]{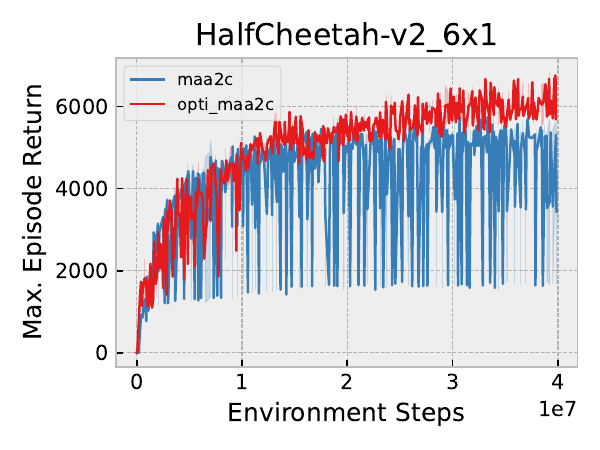}
        \label{fig:maa2c_max}
    }
    \caption{Comparisons of Optimistic MAA2C and vanilla MAA2C on \textit{HalfCheetah 6x1} task. The left figure shows the average episode returns and the right shows maximum episode returns.}
    \label{fig:maa2c_results}
\end{figure}

We also apply the proposed advantage clipping to multi-agent advantage actor-critic (MAA2C)~\cite{papoudakis2020benchmarking}, dubbed OptiMAA2C. Figure~\ref{fig:maa2c_results} demonstrates that OptiMAA2C improves MAA2C similarly to how OptiMAPPO improves MAPPO.

\section{Comparison with NA-MAPPO on \textit{MA-MuJoCo}}\label{sec:namppo}
\begin{figure}[H]
    \centering
    \subfigure[Average episode returns]{
        \includegraphics[width=0.7\textwidth]{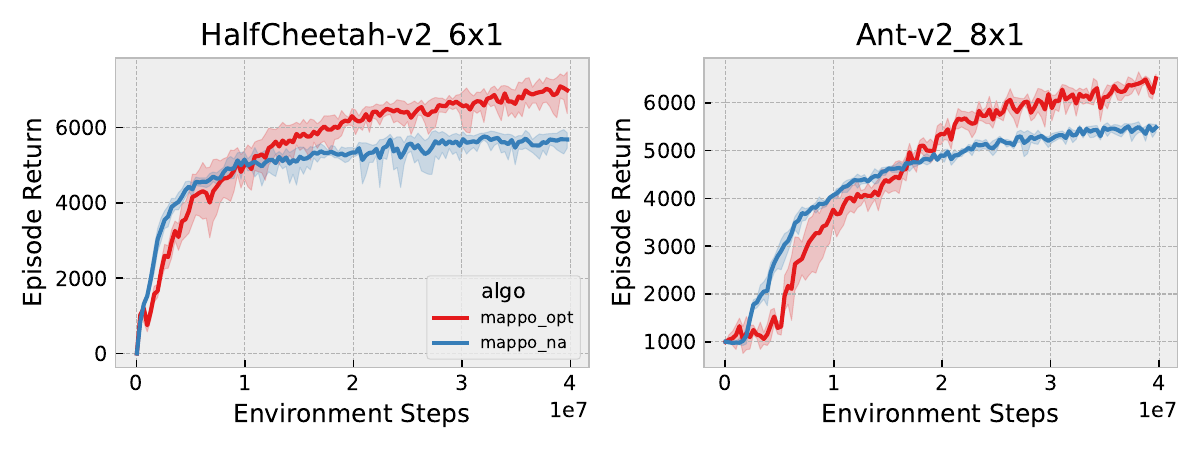}
        \label{fig:namappo_ave}
    }
    \subfigure[Maximum episode returns]{
        \includegraphics[width=0.7\textwidth]{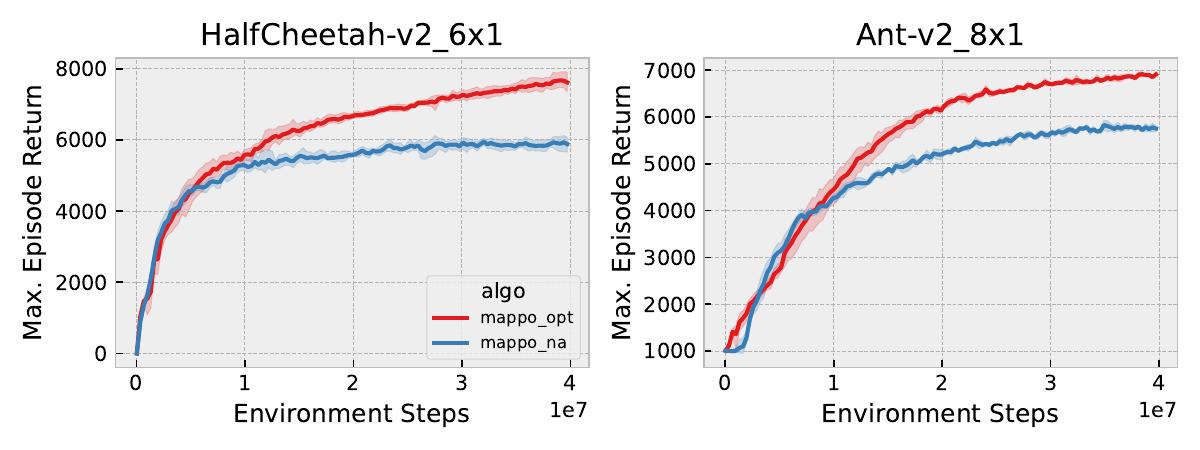}
        \label{fig:namappo_max}
    }
    \caption{Comparisons with NA-MAPPO (\textit{mappo\_na} in the figure) on \textit{HalfCheetah 6x1} and \textit{Ant 8x1} tasks. The top figure shows the average episode returns and the bottom shows maximum episode returns.}
    \label{fig:namappo_results}
\end{figure}

NA-MAPPO~\cite{hu2021policy} modifies the advantage estimates by injecting Gaussian noise in order to enhance the exploration of MAPPO. We compare our method with NA-MAPPO on two \textit{MA-MuJoCo} tasks, \textit{HalfCheetah 6x1} and \textit{Ant 8x1}. The result in Figure~\ref{fig:namappo_results} shows that our method consistently outperforms NA-MAPPO. 

\section{Full Results on \textit{MA-MuJoCo}}\label{sec:mujoco_full}

\begin{figure}[ht]
\centering
\includegraphics[width=0.96\textwidth, ]{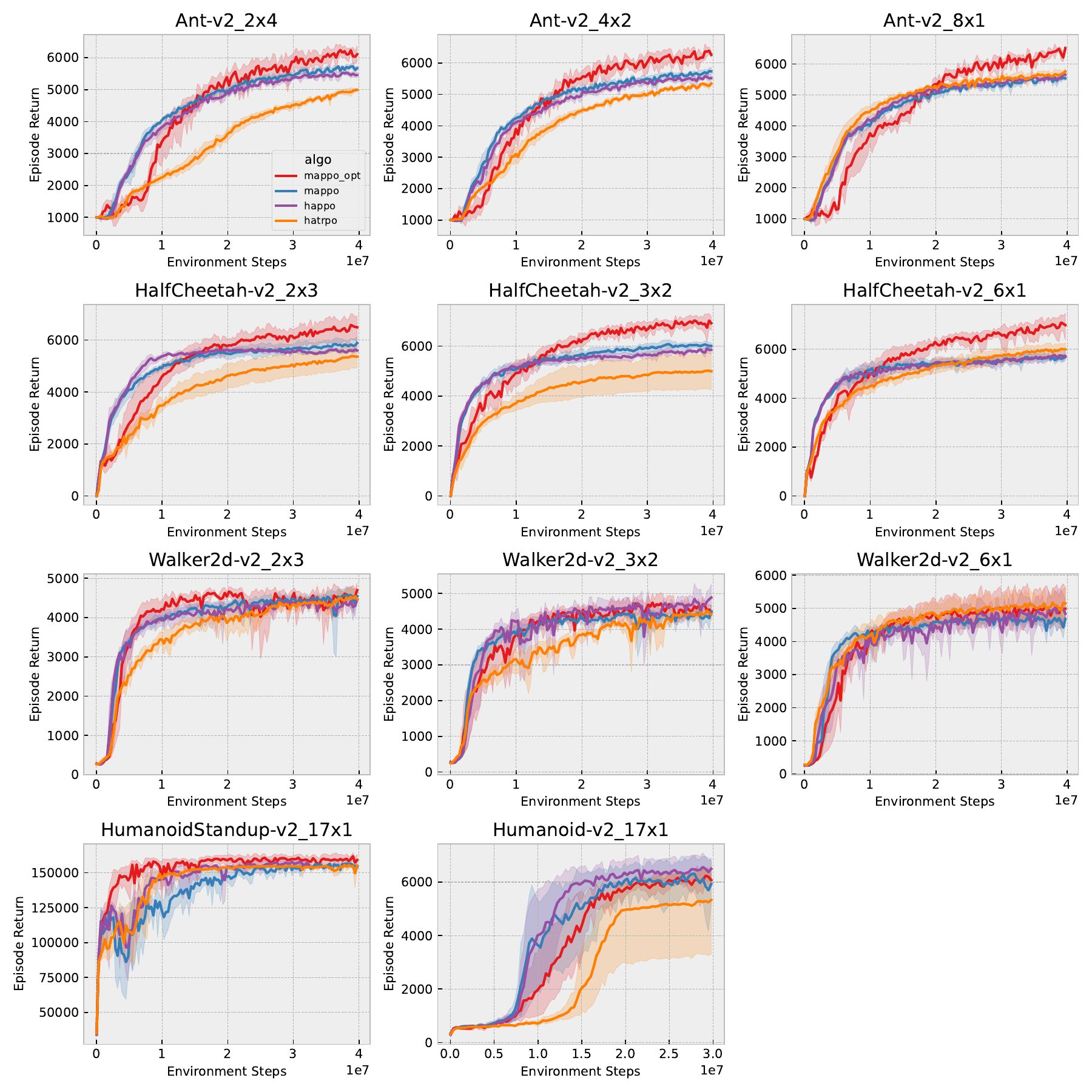}
% \vspace{-1em}
\caption{Comparisons of average episode returns on \textit{MA-MuJoCo} tasks. OptiMAPPO (\textit{mappo\_opt} in the figures) converges to a better joint policy in most tasks, especially the \textit{Ant} and \textit{HalfCheetah} tasks. We plot the mean across 5 random seeds, and the shaded areas denote 95\% confidence intervals.}
\label{fig: mujoco_full}
\end{figure}

\begin{figure}[ht]
\centering
\includegraphics[width=\textwidth, ]{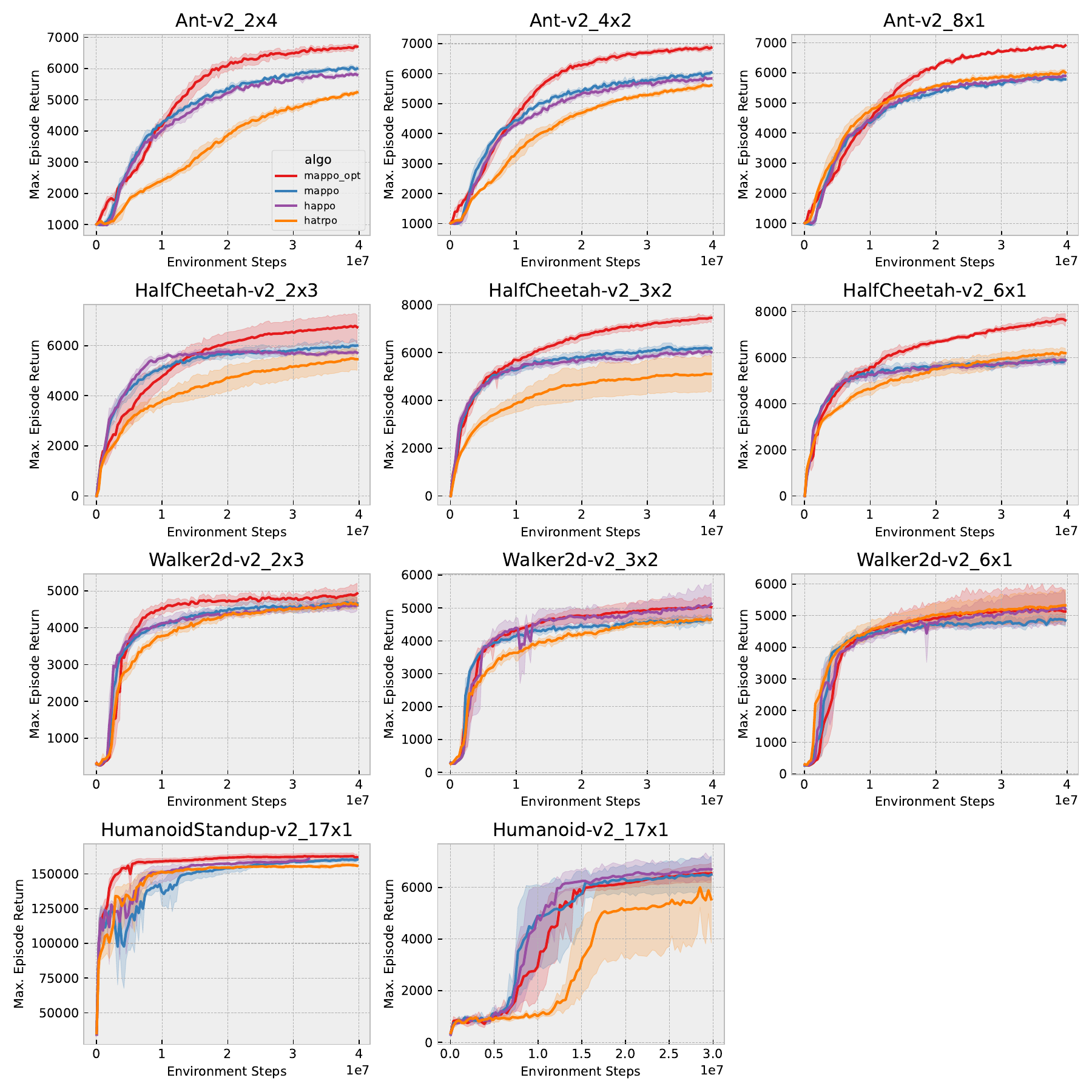}

% \vspace{-1.2em}
\caption{The maximum episode returns on \textit{MA-MuJoCo} tasks, where our method OptiMAPPO (\textit{mappo\_opt} in the figures) outperforms the strong baselines on most tasks with clear margins.}

\label{fig: mujoco_max}
\end{figure}

We show the average and maximum return of $100$ evaluation episodes during training on all $11$ \textit{MA-MuJoCo} tasks in Figure~\ref{fig: mujoco_full} and Figure~\ref{fig: mujoco_max}, which shows our method outperforms the baselines on most tasks and matches the rest. With respect to the maximum episode return in Figure~\ref{fig: mujoco_max}, our algorithm demonstrates clearer margins over the baselines.

%%%%%%%%%%%%%%%%%%%%%%%%%%%%%%%%%%%%%%%%%%%%%%%%%%%%%%%%%%%%%%%%%%%%%%%%%%%%%%%
%%%%%%%%%%%%%%%%%%%%%%%%%%%%%%%%%%%%%%%%%%%%%%%%%%%%%%%%%%%%%%%%%%%%%%%%%%%%%%%

\end{document}